\pdfoutput=1
\documentclass[10pt,twocolumn,letterpaper]{article}

\usepackage{iccv}
\usepackage[utf8]{inputenc}
\usepackage{times}
\usepackage{epsfig}

\usepackage{amsmath,amsfonts,bm}

\def\eqref#1{equation~\ref{#1}}

\def\1{\bm{1}}

\def\vb{{\bm{b}}}

\def\vk{{\bm{k}}}

\def\vp{{\bm{p}}}

\def\vr{{\bm{r}}}

\def\vv{{\bm{v}}}

\def\vx{{\bm{x}}}
\def\vy{{\bm{y}}}
\def\vz{{\bm{z}}}

\DeclareMathAlphabet{\mathsfit}{\encodingdefault}{\sfdefault}{m}{sl}
\SetMathAlphabet{\mathsfit}{bold}{\encodingdefault}{\sfdefault}{bx}{n}

\usepackage{graphicx}
\usepackage{hyperref}
\usepackage{url}
\usepackage{amsmath,amssymb} 
\usepackage{color}
\usepackage{subcaption}
\usepackage{adjustbox}
\usepackage{tabu}
\usepackage{textcomp}
\usepackage{tabularx}
\usepackage{placeins}
\usepackage[font={small}]{caption}
\usepackage{amsthm}
\usepackage{amsthm}
\usepackage[]{algorithm2e}

\theoremstyle{plain}

\newtheorem{theorem}{Theorem}
\newtheorem{lemma}[theorem]{Lemma}

\newcommand{\fig}[1]{Figure~\ref{fig:#1}}

\newcommand{\tab}[1]{Table~\ref{tab:#1}}

\newcommand{\dpart}[2]{\frac{\partial #1}{\partial #2}}

\newcommand{\RR}{\mathbb{R}}

\makeatletter\renewcommand\paragraph{\@startsection{paragraph}{4}{\z@}
  {.5em \@plus1ex \@minus.2ex}{-.5em}{\normalfont\normalsize\bfseries}}\makeatother

\def\iccvPaperID{8848} 

\iccvfinalcopy

\ificcvfinal\fi

\hypersetup{
    colorlinks=true,
    linkcolor=blue,
    filecolor=magenta,      
    urlcolor=magenta,
}

\begin{document}

\title{Cloud Transformers: A Universal Approach To Point Cloud Processing Tasks}
\author{Kirill Mazur\textsuperscript{1} \qquad\qquad Victor Lempitsky\textsuperscript{1, 2}\thanks{VL is currently with Yandex and Skoltech.}\\
\textsuperscript{1}Samsung AI Center Moscow \quad\textsuperscript{2}Skolkovo Institute of Science and Technology (Skoltech)\\
}

\maketitle

\begin{abstract}
We present a new versatile building block for deep point cloud processing architectures that is equally suited for diverse tasks. This building block combines the ideas of spatial transformers and multi-view convolutional networks with the efficiency of standard convolutional layers in two and three-dimensional dense grids. The new block operates via multiple parallel heads, whereas each head differentiably rasterizes feature representations of individual points into a low-dimensional space, and then uses dense convolution to propagate information across points. The results of the processing of individual heads are then combined together resulting in the update of point features. Using the new block, we build architectures for both discriminative (point cloud segmentation, point cloud classification) and generative (point cloud inpainting and image-based point cloud reconstruction) tasks. The resulting architectures achieve state-of-the-art performance for these tasks, demonstrating the versatility of the new block for point cloud processing. 
\end{abstract}

\section{Introduction}
\label{sect:intro}



In this work, we consider recognition and generation tasks for point clouds, such as semantic segmentation or image-based reconstruction. Most state-of-the-art architectures for point cloud processing are derived from convolutional neural networks (ConvNets)~\cite{LeCun89} and are inspired by the success of ConvNets in image processing tasks. Such ConvNet adaptations are based on direct rasterization of point clouds onto regular grids followed by convolutional pipelines~\cite{Su15, Graham17}, as well as on generalizations of the convolutional operators to irregularly sampled data~\cite{Mao19, WangParam18} or non-rectangular grids~\cite{Klokov17, Jampani16}.

In this work, we propose a new building block (a \textit{cloud transform} block) for point cloud processing architectures that combines the ideas of ConvNets and Transformers~\cite{Vaswani17} (\fig{block}). Similarly to the (self)-attention layers within transformers, our cloud transform blocks take unordered sets of vectors as inputs, and process such input using multiple parallel \textit{heads}. For an input set element, each head computes two- or three-dimensional \textit{key} and a higher dimensional \textit{value}, and then uses the computed keys to rasterize the respective values onto a regular grid. A two- or three-dimensional convolution is then used to propagate the information across elements. The results of parallel heads are then probed at key locations and are recombined together, producing an update to element features.


\begin{figure}
    \centering
    \includegraphics[width=\columnwidth]{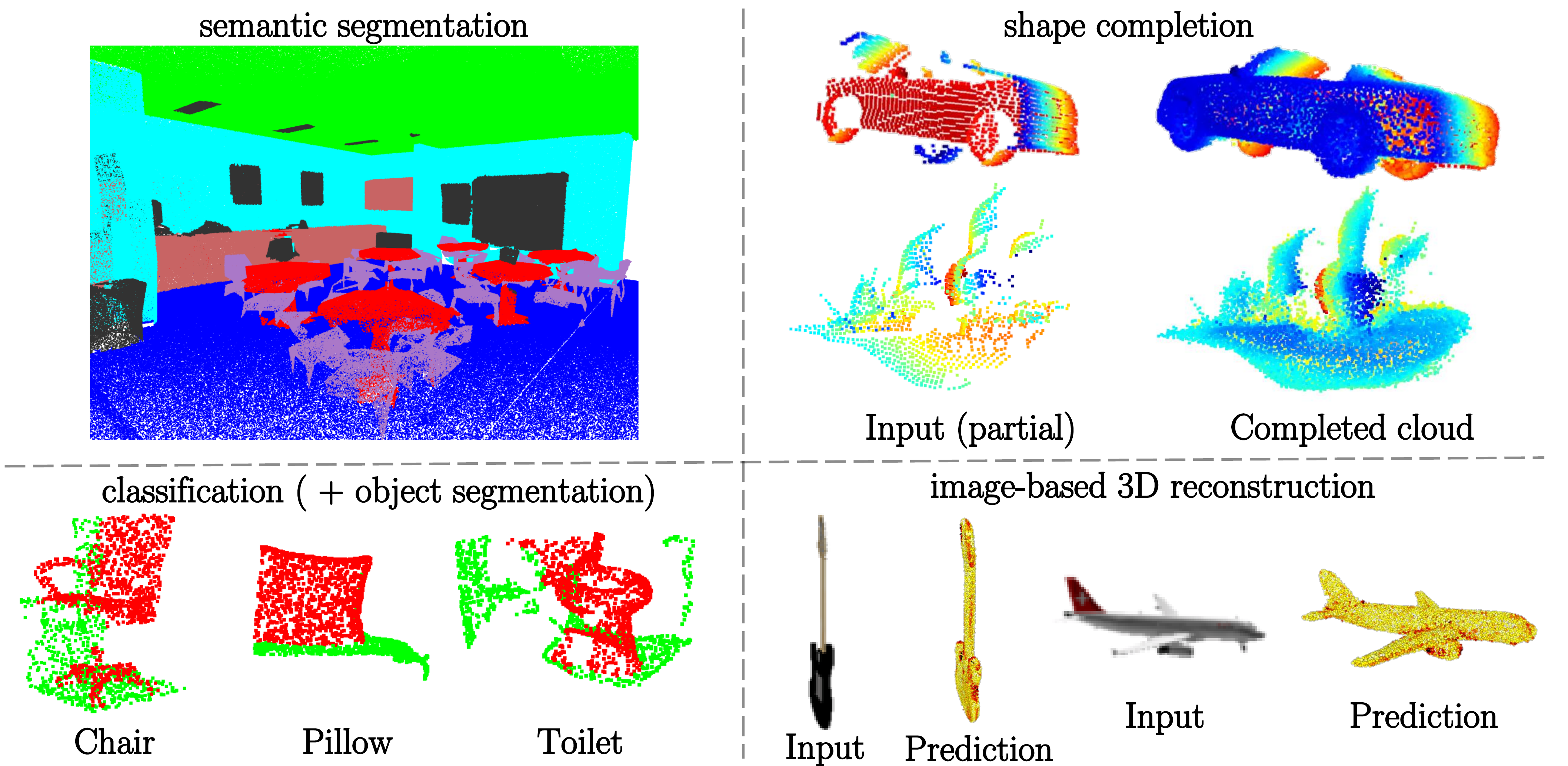}
    \caption{Sample outputs of cloud transformers across four diverse cloud processing tasks including recognition tasks (left) and generation tasks (right).\vspace{-0.5cm}}
    \label{fig:teaser}
\end{figure}

We show that multiple cloud transform blocks can be stacked sequentially and trained end-to-end, as long as special care is taken when implementing forward and backward pass through the rasterization operations. We then design \textit{cloud transformer} architectures that concatenate multiple cloud transform blocks together with task-specific 3D convolutional layers. Specifically, we design a cloud transformer for semantic segmentation (which we evaluate on the S3DIS benchmark~\cite{armeni_cvpr16}), classification (which we evaluate on the ScanObjectNN benchmark~\cite{Uy19}), point cloud inpainting (which we evaluate on ShapeNet-based benchmark~\cite{yuan2018}), and a cloud transformer for image-based geometric reconstruction (which we evaluate on a recently introduced ShapeNet-based benchmark~\cite{Tatarchenko19}). In the evaluation, the designed cloud transformers achieve state-of-the-art accuracy for semantic segmentation and point cloud completion tasks and considerably outperform state-of-the-art for image-based reconstruction and point cloud classification (\fig{teaser}). We note that such versatility is rare among previously introduced point cloud processing architectures, which can handle either recognition tasks (such as semantic segmentation, classification) or generation tasks (such as inpainting and image-based reconstruction) but usually not both.

To sum up, our key contributions and novelty are:
    \begin{itemize}
        \setlength\itemsep{0.01em}
        
        \item We propose a new approach to point cloud processing based on repeated learnable projection, rasterization and de-rasterization operations. We investgate how to make rasterizations and de-rasterizations repeatable \textit{sequentially} within the same architecture through the gradient balancing trick. Additionally, we show that aggregating rasterizations via element-wise maximum performs better than additive accumulation at least in the context of our approach.
        
        \item We propose and validate an idea of multi-head self-attention for point clouds that performs \textit{parallel} processing by rasterization and de-rasterization to separate low-dimensional grids. Additionally, we propose an idea of using both two-dimensional and three-dimensional grids in parallel with each other. 
        
        \item Based on the two ideas above, we propose architectures for semantic segmentation, classification, point cloud inpainting, and image-based reconstruction. The proposed architectures are all based on the same Cloud Transform blocks and achieve state-of-the-art performance on standard benchmarks in each case despite the diversity of tasks.
    \end{itemize}

\section{Related work}
\label{sect:related}


A number of work use rasterizations of the point cloud over regular 3D grids~\cite{Maturana15,Graham17,Moon2017V2VPoseNetVP,Mao19}, where each point is rasterized at its original position within the point cloud. Multi-view ConvNets~\cite{Su15} project point clouds to multiple predefined 2D views. The approaches that use splat convolutions on permutohedral grid convolutions~\cite{Kiefel15, Su18} are perhaps most similar to ours (and have been an inspiration to us), as they also interleave rasterization (\textit{splatting}), (permutohedral) convolution, and probing (\textit{slicing}). In contrast to all above-mentioned works, which use initial positions or data-independent projections of points for rasterizations, our architectures use diverse data-dependent projections.

A dynamic graph ConvNet (DGCNN) architecture~\cite{Wang19} uses a graph ConvNet. The graph is computed from spatial positions of the points that are modified in a data-dependent way within the architecture. In their case, the loss can not be backpropagated through the graph node position estimation since the spatial graph construction is non-differentiable. In contrast, our approach is based on regular grid convolutions and includes the backpropagation through position estimation (key computation). We also note that differentiable point cloud projection onto a 2D grid (from 3D space) has been used in~\cite{Insafutdinov18} though in a different way and for a different purpose than in our case.

Our approach is also related to \textit{spatial transformers}~\cite{Jaderberg15}, i.e.\ neural blocks that warp signals on regular grids through data-dependent parametric warping and bilinear sampling. Our blocks also use bilinear sampling in the end of each head processing. Inspired by spatial transformers, \cite{Wang19a} investigate how data-independent and data-dependent deformations of the original point clouds can be used to boost the performance of several recognition architectures including DGCNN~\cite{Simonovsky17}, SplatNet~\cite{Su18}, and VoxelNet~\cite{Zhou17}. Similarly to \cite{Wang19} and unlike \cite{Jaderberg15}, \cite{Wang19a} do not propagate the loss fully through deformation computation (in the case of data-dependent deformations). Compared to \cite{Wang19a}, our architectures employ regular 2D and 3D convolutions, can handle both recognition and generative tasks (the latter not considered in \cite{Wang19a}), and are trained with gradient propagation through key position computation. Our work is also related to \cite{Liu2019, Tang2020}, as our method is also based on rasterizations and de-rasterizations. However, these methods are not applicable for data-dependent transformations of rasterizaton positions and, therefore, cannot handle generative tasks. Additionally, we employ a different method for rasterizations via element-wise maximum, instead of averaging. 


Concurrently with us, the work \cite{Zhao2021} also adapted transformers to the point cloud domain. Their Point Transformer architecture handles the large size of point clouds by restricting self-attention blocks to considering nearest neighbors of individual points, and otherwise follows the original transformer architecture~\cite{Vaswani17} closely. Our work replaces explicit self-attention mechanism with the combination of rasterization and convolution, and thus shares less similarity with~\cite{Vaswani17} and more similarity with works that rely on ConvNets. Point Transformer significantly outperforms state-of-the-art for discriminative tasks, and outperforms our architecture on a semantic segmentation benchmark. At the same time, the adaptation of their approach to generative tasks is not immediately obvious.

\section{Method}
\label{sect:method}

\paragraph{Overview.} We describe our approach in a \textbf{bottom-up} way. In Section~\ref{sect:cloud_transform}, we introduce the basic building operation of our processing pipeline that we call \textit{cloud transform}. In Section~\ref{sect:multihead}, we discuss how cloud transforms can be assembled in a parallel fashion into blocks (which we call \textit{multi-head processing blocks}). In Section~\ref{sect:cascaded_mhct}, we discuss sequential stacking of multi-head processing blocks into bigger blocks (called \textit{cascaded multi-head processing blocks}) that cycle through different spatial resolution and different numbers of feature channels. Finally, in Section~\ref{sect:cloud_transformers}, we introduce the architectures (that we call \textit{cloud transformers}) build from cascaded multi-head processing blocks for four different point cloud processing tasks.



\subsection{Cloud Transform}
\label{sect:cloud_transform}

\begin{figure}
\centering
  \includegraphics[width=\the\columnwidth]{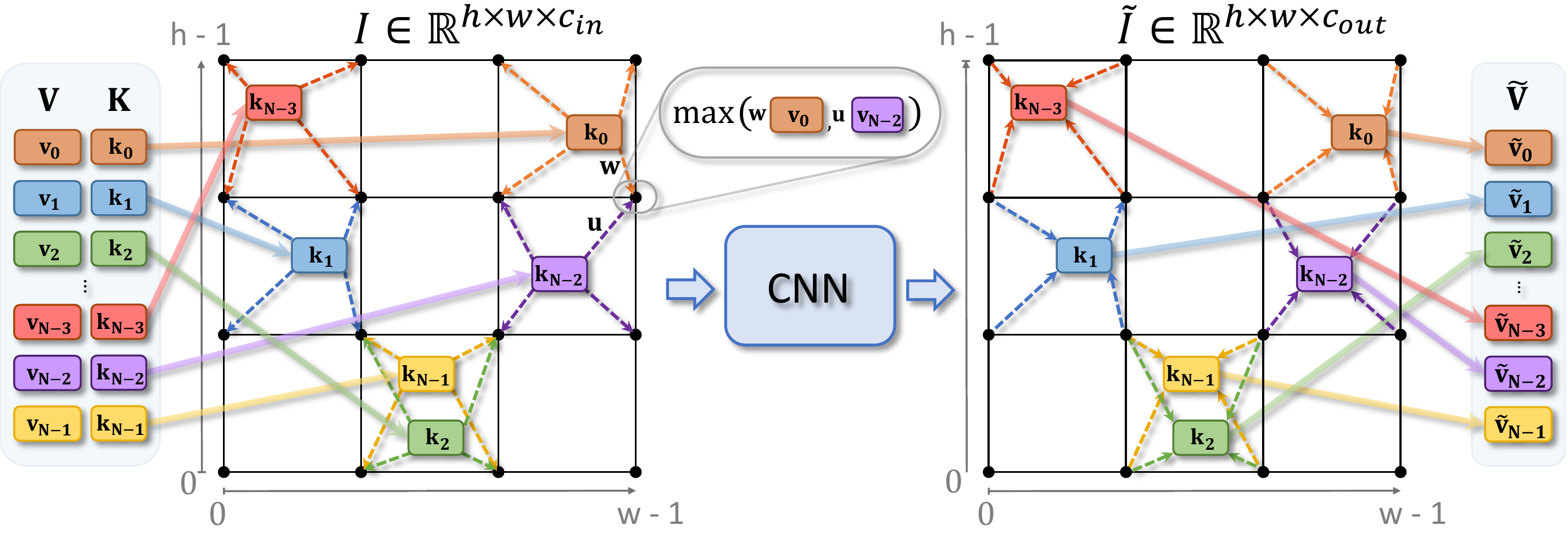}
  \vspace{-5ex}
\caption{The cloud transform consists of rasterization (left) and de-rasterization(right) steps, with the convolutional part in between. It projects the high-dimensional point cloud onto low-dimensional (two-dimensional in this case) grid, applies convolutional processing, and lifts the result back to the high-dimensional space. Electronic zoom-in recommended.}
\label{fig:cloud_transform}
\vspace{-4ex}
\end{figure}
\vspace{-1ex}
The cloud transform contains three steps: rasterization, convolution, and de-rasterization.
\paragraph{Rasterization step.} To rasterize each point $\vx_i$, we predict the \emph{value} $\vv_i \in \RR^{c_{in}}$ and the \emph{key} $\vk_i \in [0, 1]^2$. These two vectors stand for \emph{what} to rasterize and \emph{where} to rasterize respectively.

The cloud transform (\fig{cloud_transform}) takes as an input an unordered set (to which we further refer as point cloud) $X = \{ \vx_1, \dots , \vx_N \mid \vx_i \in \RR^f \}$, whose elements are vectors  $\vx_i \in \RR^{f}$ of a potentially high dimension $f$. The Cloud Transform $\mathcal{T}(X)$ maps such input into a new $g$-dimensional point cloud $Y \in \RR^{N \times g}$ of the same size $N$. Additionally, our layer uses an input point cloud positions $P = \{ \vp_1, \dots , \vp_N \mid \vp_i \in \RR^3 \}$.



The cloud transform first applies a learnable projection $\mathcal{P}_{2}$ (further called \emph{rasterization}), which generates a two-dimensional feature map with $c_{in}$ channels, i.e.\ $\mathcal{P}_{2} \colon X \mapsto I \in \RR^{ w \times w \times c_{in}}$. In a volumetric setting, the cloud transform starts with a learnable projection $\mathcal{P}_{3}$, which generates a three-dimensional volumetric feature map, i.e.\  $\mathcal{P}_{3} \colon X \mapsto I \in \RR^{w \times w \times w \times c_{in}}$. In both cases, $w$ stands for the spatial resolution of the grid, while $c_{in}$ stands for the number of input channels. 

 Once an irregular point cloud $X$ is projected onto a regular feature map, the cloud transform applies a single convolution or a more complex combination of convolutional operations. We denote the result of these convolutional layers as $\tilde{I} \in \RR^{w \times w \times c_{out}}$ ($\tilde{I} \in \RR^{w \times w \times w \times c_{out}} $ in the volumetric case). Note that we expect $\tilde{I}$ to be of the same spatial size as $I$. However, the channel dimension of $\tilde{I}$ might be changed from $c_{in}$ to $c_{out}$.  

 The last step of our Cloud Transform operation is \emph{de-rasterization} (also called \emph{slicing}) $\tilde{\mathcal{P}} \colon \tilde{I} \to \widetilde{V}$ from the processed feature map $\tilde{I}$ into a new \emph{transformed} values $\widetilde{V} \in \RR^{N \times c_{out}}$. Note, that cloud transform passes information from $\vx_i$ to $\vx_j$ as long as these two points have been projected to sufficiently close positions. Thus, the cloud transform can be seen as a variant of self-attention layer with adaptive sparse attention mechanism. Below, for the sake of simplicity, we detail the steps of the cloud transform  for a two-dimensional feature map case. The volumetric case is completely analogous.


Our method allows to directly predict keys via linear layer and stack these layers into deep architectures. However, in our experiments transforming positions via matrix multiplication (i.e predicting it with a point-wise MLP from $\vx_i$) results in suboptimal performance. A better solution predicts deep residuals $d(\vx_i)$ to the input positions $\vp_i$ with a single layer peceptron $d$ and apply a learnable transformation $T \in SE(3)$ afterwards (the transform $T$ becomes the parameter of the layer). The keys are thus computed as:
\vspace{-1ex}
\begin{equation}
    \label{eq:pred_keys}
    \vk_i = T \left(\vp_i + d(\vx_i) \right)
\vspace{-1ex}
\end{equation}
In the case of two-dimensional heads, we project $\vk_i$ to the $z = 0$ plane (omitting the third coordinate). Finally, we apply the sigmoid activation to the keys $\vk_i$ ensuring they are positioned between zero and one. As for the values $\vv_i$ prediction, we use a single affine layer with the output dimension equal to $c_{in}$, followed by the normalization layer.

Depending on the architecture, the normalization layer can be batch normalization~\cite{pmlr-v37-ioffe15}, instance normalization~\cite{Ulyanov2017ImprovedTN} or adaptive instance normalization~\cite{Huang17}.





We then rasterize the value $\vv_i \in \RR^c_{in}$ onto the grid $I = \RR^{w \times w \times c_{in}}$ using the predicted key $\vk_i$ as a position. Specifically, $\vk_i = (k^0_i, k^1_i) \in [0, 1]^2$ may be interpreted as a relative coordinate inside the spatial grid of $I$. Thus, the position defined by $\vk_i$ falls into the enclosing integer cell $(h_0, w_0)$, $(h_0, w_1)$, $(h_1, w_0)$, $(h_1, w_1)$, where $h_0{=}\lfloor{}(w{-}1){\cdot} k_i^0 \rfloor{}$, $h_1{=}\lceil{}(w{-}1){\cdot} k_i^0\rceil{}$,$w_0{=}\lfloor{}(w{-}1){\cdot} k_i^1 \rfloor{}$, 
$w_1{=}\lceil{}(w{-}1){\cdot} k_i^1 \rceil{}$. The value $\vv_i$ is then rasterized into four neighbouring feature map pixels $I[h_0, w_0]$, $I[h_0, w_1]$, $I[h_1, w_0]$, $I[h_1, w_1] \in \RR^{c_{in}}$ via bilinear assignment. In more detail, we compute bilinear weights $\vb_i = (b_i^{00}, b_i^{01}, b_i^{10}, b_i^{11})$  of the key $\vk_i$ with respect to the cell it falls to. The bilinear weights are then used to update the feature map $I$ at corresponding locations:
\vspace{-2ex} \begin{align} 
\begin{split}
\label{eq:splat}
    I[h_0, w_0] & \leftarrow  \operatorname{max}(I[h_0, w_0], b_i^{00} \vv_i) \\ 
    I[h_0, w_1] & \leftarrow  \operatorname{max}(I[h_0, w_1], b_i^{01} \vv_i) \\ 
    I[h_1, w_1] & \leftarrow  \operatorname{max}(I[h_1, w_0], b_i^{10} \vv_i) \\ 
    I[h_1, w_1] & \leftarrow  \operatorname{max}(I[h_1, w_1], b_i^{11} \vv_i)  
\end{split}
\end{align} \vspace{-0.5ex}
The feature map $I$ is initialized with zeros, and the rasterization is repeated for every $\vx_i \in X$, $i \in 1..N$, aggregating rasterized results via element-wise maximum at respective cells of the feature map $I$. While the choice of the maximum aggregator may seem unnatural compared to average or sum, we have found that it boosts the performance substantially.
\vspace{-1ex}
\paragraph{Convolution step.} As discussed above, after rasterization, we transform the feature map $I$ into $\tilde{I}$ with any convolutional architecture that preserves the spatial resolution. In practice, we use a single convolutional layer that keeps the number of channels unchanged. 
\vspace{-1ex}
\paragraph{De-rasterization step.} As the last step, we perform the \emph{de-rasterization} transform $\tilde{\mathcal{P}}_{2} \colon \tilde{I} \to \tilde{V}$ produces the transformed feature cloud $Y$ using standard bilinear grid sampling operation. Thus, the transformed values $\tilde{I}[h_0, w_0]$, $\tilde{I}[h_0, w_1]$, $\tilde{I}[h_1, w_0]$, $\tilde{I}[h_1, w_1] \in \RR^{c_{out}}$ of the feature map are combined with bilinear weights $\vb_i$ into the transformed value vector $\tilde{\vv}_{i}$.

We apply the normalization layer, and the ReLU nonlinearity to the result of de-rasterization step, and further map each value from $c_{out}$ dimensions back to $g$ dimensions ($g{=}512$ unless noted otherwise) using a learnable affine transform.



\paragraph{Backpropagation through Cloud Transform.} We have found that learning architectures with multiple sequentially-stacked Cloud Transform blocks via back-propagation~\cite{Rumelhart86} is highly unstable, as the gradients explode during the backward step. 
An ideal assumption on gradient variance during the back-propagation is to preserve its scale throughout the network~\cite{Glorot10, He15}. In the supplementary we demonstrate that a na\"ive verison of Cloud Transform block could not satisfy this assumption and suggest a \textit{gradient balancing} trick to solve this issue. In our case, the instability can be tracked to the gradient of the bilinear weights $\vb$ w.r.t. the key $\vk$ at the rasterization and de-rasterization steps. According to the chain rule, the gradients' variance is multiplied by $w$ during backpropagation through the keys, which results in the exponential resulting gradient variance (w.r.t. depth).


\paragraph{Gradient balancing trick} Based on observation above, during back-propagation of $\mathcal{L}$ through keys, we simply divide the partial derivatives w.r.t.\ both coordinates of $\vk_i$ by $w$, i.e.\ we apply:

\begin{equation}
      \dpart{\mathcal{L}}{\vk_i} \leftarrow \frac{1}{w}\dpart{\mathcal{L}}{\vk_i}\,.
\end{equation} 

We have found that this \textit{gradient balancing} trick is sufficient to enable the learning of deep architectures containing multiple layers with cloud transforms.

\subsection{Multi-Headed Cloud Transform block}
\label{sect:multihead}

\begin{figure}
\centering
  \includegraphics[width=\the\columnwidth]{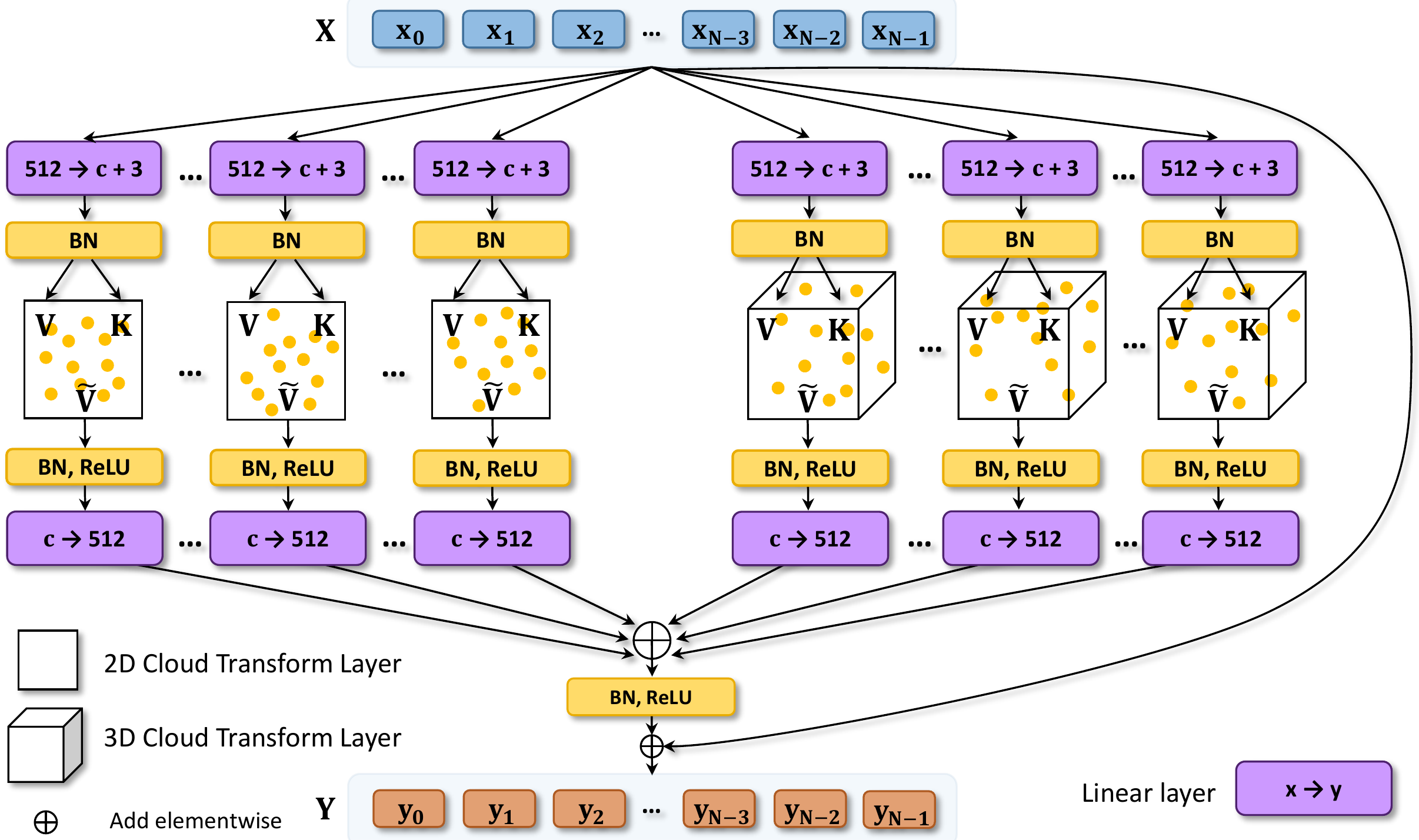}
\caption{Our building block has several planar heads and several volumetric heads operating in parallel. Each head is a cloud transform, using a two-dimensional or a three-dimensional grid for rasterization, followed by convolutional operations, and de-rasterization (differentiable sampling). Electronic zoom-in recommended.}
\label{fig:block}
\vspace{-3ex}
\end{figure}

The rasterization and de-rasterization operation may lead to the information loss due to the limited number of nodes in two dimensional and three dimensional lattices (we use up to $w{=}128$ for two-dimensional grids and up to $w{=}32$ for three-dimensional grids). We therefore build our architectures from blocks that combine multiple cloud transforms operating in parallel. This is reminiscent of both the multiple self-attention head in the Transformer architecture~\cite{Vaswani17} and the multi-view convolutional networks~\cite{Su15}. Following~\cite{Vaswani17}, we call each of the parallel cloud transform modules a \textit{head} and thus consider a multi-head architecture. Each head predicts keys and values independently, and may use its own spatial resolution $w$. In fact, two-dimensional and three-dimensional heads can operate in parallel. We note that the use of one-dimensional or higher-dimensional (e.g.~four-dimensional) heads is also possible. One-dimensional grids however inevitably results in very strong conflation of the data, while higher-dimensional grids are computationally heavy and convolutions on them are not well supported. We therefore focus on two- and three-dimensional heads. 

The results of the parallel heads for each point $i$ are summed together, so that the resulting \textit{multi-head cloud transform (MHCT) block} (\fig{block}) still maps each input vector $\vx_i$ to a $g$-dimensional vector $\bm{y}_i$. We add another normalization layer and ReLU nonlinearity after the results of the heads are summed, and complete the block with the residual skip connection from the start to the end.\cite{He15}.
We note that the multi-head cloud transform block also resembles the Inception block~\cite{Szegedy15a}, which uses heterogeneous parallel convolutions, as well as the blocks of the ResNeXt networks~\cite{Xie16}, which use grouped convolutions with small number of channels in each group.

In this work we use MHCT blocks with $16$ two-dimensional heads and $16$ three-dimensional heads. 

\begin{figure}[t]
\centering \includegraphics[width=0.9\columnwidth]{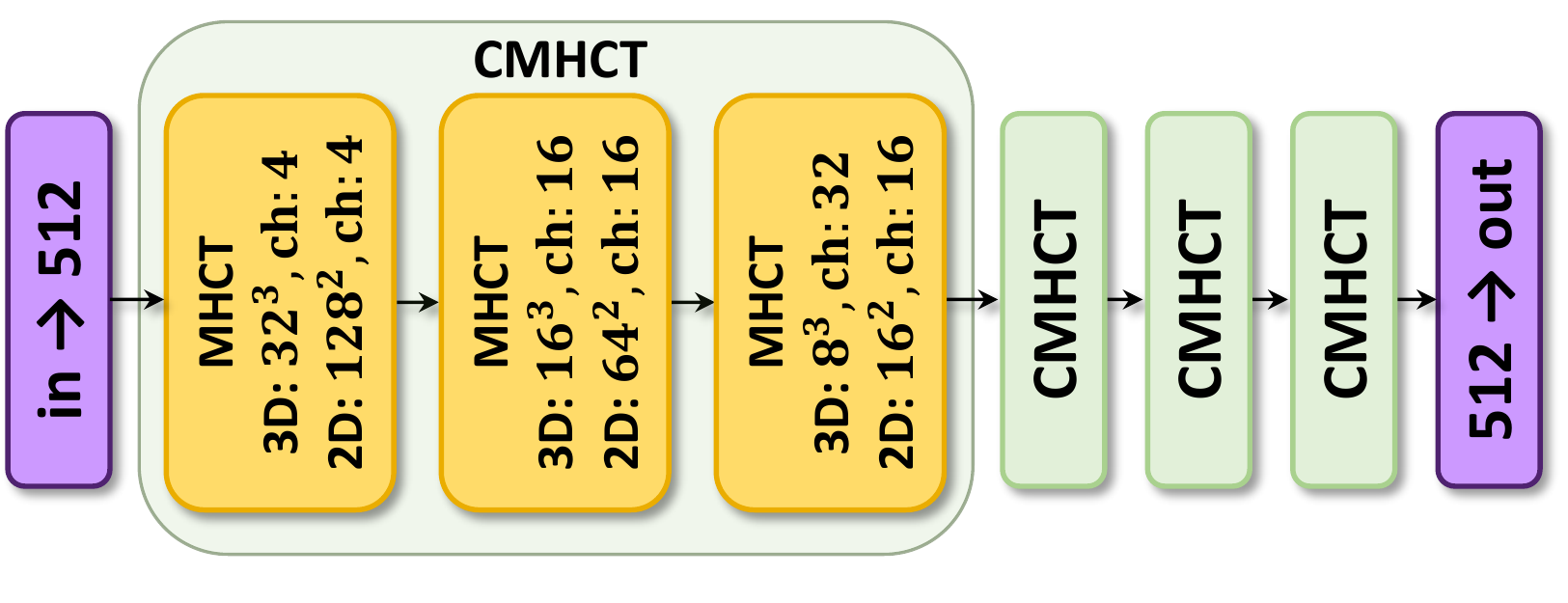} \vspace{-0.5cm}
\caption{The architecture used for semantic segmentation is based on (1) a single point-wise layer, (2) four standard cascaded multi-headed cloud transform blocks, followed by (3) a final point-wise MLP.}
\label{fig:segmenter}
\vspace{-3ex}
\end{figure}

\subsection{Cascaded Blocks}
\label{sect:cascaded_mhct}


Our network does not use pooling and upsampling operations directly on points, as is done in most of the point cloud processing networks (e.g \cite{Thomas2019KPConvFA}, \cite{Liu2020}). Instead, we employ feature maps of different spatial sizes as a way of increasing or decreasing the receptive field. Specifically, following the pattern introduced in \cite{Simonyan15}, we propose to stack three MHCT blocks sequentially into \textit{Cascaded Multi-Headed Cloud Transform Block (CMHCT)}, decreasing spatial dimension, while increasing the number of channels. In practice, we set the spatial and channels dimensions as in the \fig{segmenter}, yellow blocks. 

The order of these three MHCT blocks in CMHCT block is as on the figure. The CMHCT blocks can then be stacked sequentially.

\subsection{Cloud Transformers}
\label{sect:cloud_transformers}
We now discuss the architectures that can be constructed from CMHCT blocks for specific point cloud processing tasks. We note that while the different nature of tasks requires different architectures, we strove to keep these architectures as similar as possible. Most importantly, all proposed architectures are build from CMHCT blocks that are built out of MHCT blocks that are based on Cloud Transforms.

\paragraph{Semantic segmentation.} The semantic segmentation cloud transformer (\fig{segmenter}) consists of an initial one-layer perceptron, which is applied to each point independently and transforms its 3D coordinates and 3D color features to an $f$-dimensional vector ($f{=}512$). Afterwards, we apply four cascaded multi-headed cloud transform layers with default setting. And then conclude the architecture with a two-layer shared perceptron that maps the features of each point to the logits of segmentation classes. All normalization layers in the architectures are BatchNorm layers~\cite{pmlr-v37-ioffe15}. The architecture has $9.6$M parameters and is trained with the cross-entropy loss.

\begin{figure}[t]
\centering \includegraphics[width=\the\columnwidth]{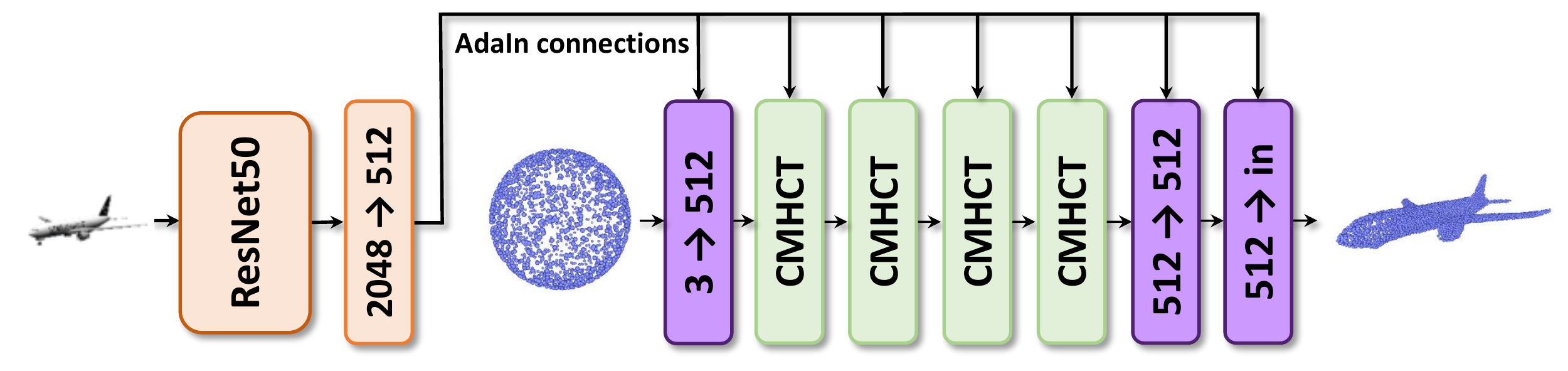} \vspace{-0.7cm}
\caption{The architecture used for image-based reconstruction consists of a convolutional style encoder (1) and a generator (2). The generator is built of a linear layer, followed by four cascaded multi-headed cloud transform blocks (green), conditioned via adaptive instance normalizations on the style vector produced by the encoder (orange). The input to the generator is sampled from a uniform sphere $S^2$.}
\label{fig:generator}
\vspace{-3ex}
\end{figure}

\paragraph{Point cloud generation.} To create the architecture that generates point cloud, we stack four CMHCT blocks sequentially, followed by a point-wise two-layer perceptron. Finally, we add a $tanh$ non-linearity to produce 3D points coordinates. The input point cloud is sampled from a uniform 3D distribution on the unit sphere $S^2$ and then passed through point-wise linear layer, mapping each feature to $f{=}512$ dimensions. 
To solve the image-based geometry reconstruction task (recovering point clouds from images), we use adaptive instance normalization (AdaIN) layers~\cite{Huang17} in the MHCT blocks. We create image encoder with ResNet-50 architecture~\cite{He15} (pretrained on ImageNet~\cite{ILSVRC15}). The output of the encoder is a $512$-dimensional vector, which is transformed into AdaIN coefficients via affine layer (\fig{generator}). The architecture is trained with approximate earth mover distance (EMD) loss \cite{Liu2019m}. Note, our generator architecture highly resembles our segmenter, apart from the normalization method. 

\begin{figure}[t]
\centering \includegraphics[width=0.9\columnwidth]{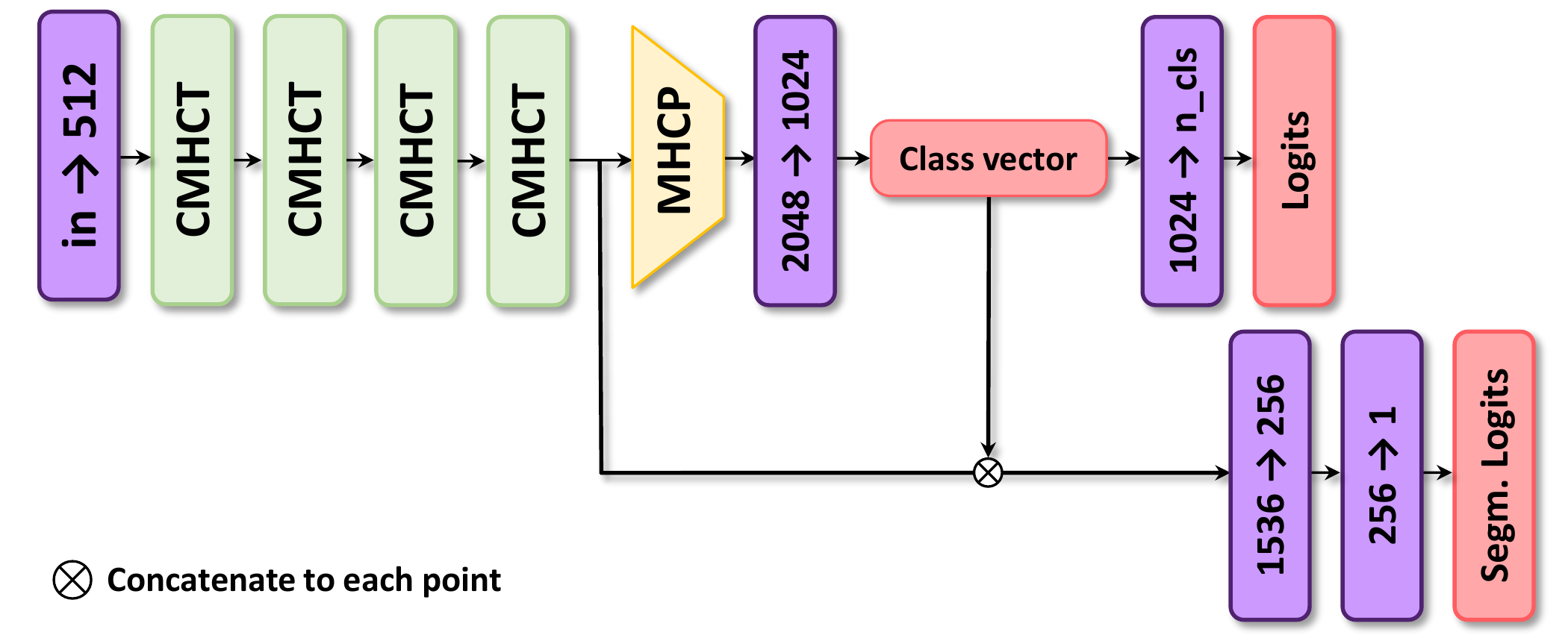} \vspace{-0.4cm}
\caption{Cloud Transformer designed for classification. Primarily, the input point cloud is processed with a Cloud Transformer backbone, as in the segmentation setting. Afterward, we apply a Multi-headed Cloud Pooling (yellow), followed by a fully-connected layer to produce a classification vector (red). We also use a separate branch (bottom) for the background mask prediction as is common for the ScanObjectNN benchmark.}
\label{fig:classifier}
\vspace{-3ex}
\end{figure}

\paragraph{Point cloud classification}
As in the segmentation model, we apply a ``backbone'' of a leading linear layer and four CMHCT blocks first (\fig{cloud_transform}). To solve a classification task, we introduce a multi-headed pooing layer. Similarly to the regular MHCT layer, this layer performs multiple rasterizations onto 2D and 3D feature maps of spatial size $8$ and $16$ respectively. The channel dimension of each head is $32$ for three-dimensional heads and $16$ for two-dimensional heads. Afterward, the resulting feature maps are processed with three standard convolutional residual blocks, each interchanged with a max-pooling (see the supplementary material for the exact architecture). The resulting vectors are aggregated across the heads via concatenation and processed with a dense layer to form a final classification vector $\vk_{class}$ of dimension $1024$. Following the original paper \cite{Uy19}, we also predict an object's instance mask, using the point-wise features predicted with from the features extracted by the backbone and the class vector $\vk_{class}$. 

We train our architecture as in \cite{Uy19} with the two cross-entropy (CE) loss terms. The first one is a CE loss $\lambda_{\text{class}}$ on object classification and the latter one is a point-wise CE loss $\lambda_{\text{seg}}$ on foreground segmentation. The final loss is set to be $\lambda_{\text{full}} \colon = 0.5 \cdot  \lambda_{\text{class}} + 0.5 \cdot  \lambda_{\text{seg}}$.
\paragraph{Point cloud inpainting}
\begin{figure}[t]
\includegraphics[width=\the\columnwidth]{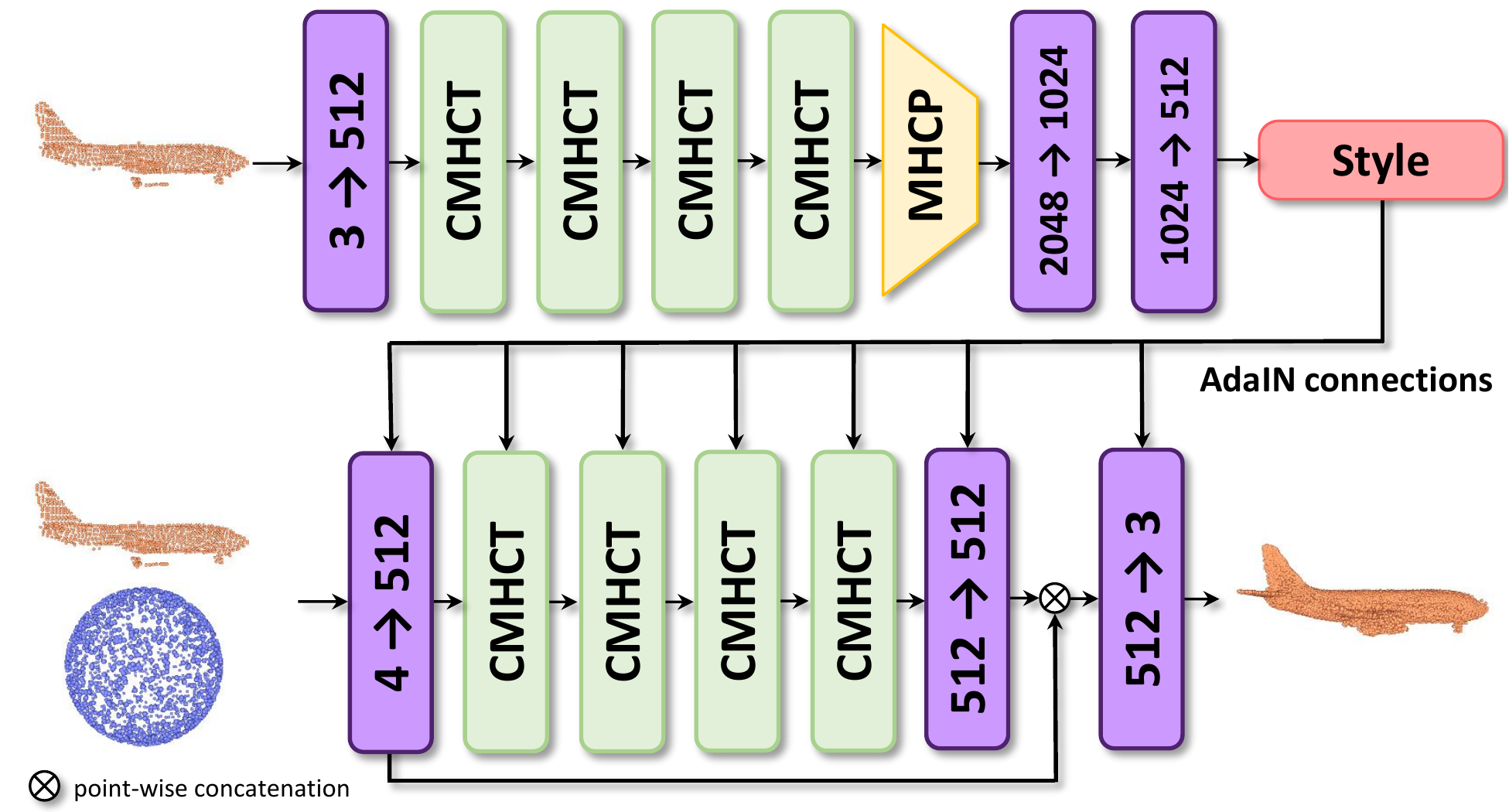} \vspace{-0.7cm}
\caption{Our Cloud Transformer Inpainter differs from our single-view reconstruction model in two ways. First, the encoder part is taken from the classifier model, producing a style vector for adaptive normalization layers of the generator. Secondly, the input to the generator consists of the partial point cloud as wekk as of the points sampled from the unit sphere $S^2$ for ``unknown'' points.}
\vspace{-3ex}
\label{fig:inpainter}
\end{figure}
We also present a model for point cloud inpainting (completion). Given a partial point cloud, the goal is to infer the full shape. Our architecture for this task strongly resembles the one used for image-based reconstruction, apart from the encoder and the generator input. Namely, we apply a point cloud encoder to obtain a style vector. Our encoder is the Cloud Transformer model introduced above for classification. It extracts a vector of dimension $512$.

To account for the input point cloud geometry, we use two sets of points in the generator branch input. First, we put in the input point cloud, thus giving the architecture opportunity to transfer its points to the output. Second, to account for the novel parts, we again sample random points from a unit sphere $S^2$. We augment the feature of each input point with a binary variable indicating whether is is sampled from the incomplete scan or from the sphere. As in the case of the image-based reconstruction, the generator branch is conditioned on the encoded vector via AdaIN connections. The model is trained with a sum of approximate EMD \cite{Liu2019m} and Chamfer Distance \cite{Fan2017} losses. Afterwards, we fine-tine it on Chamfer Distance loss.






\section{Experiments}
We compare the performance of our approach to the state-of-the-art on the four considered tasks, using established benchmarks and metrics associated with those benchmarks. We then perform a short ablation study. In the \textbf{supplementary materials}, we provide visualizations of the cloud transformer operations and results, and the readers are encouraged to check the material to gain better intuition about cloud transformers.

\label{sect:experiments}



\paragraph{Point Cloud Classification.} We evaluate our Cloud Transformer for classification on a real-world dataset ScanObjectNN~\cite{Uy19}. The dataset consists of $2902$ unique objects obtained from ScanNet~\cite{Dai2017} scenes. The objects are categorized into $15$~classes. Each object is obtained via cutting it from the scene using the ground truth bounding box. Note that the resulting cut may include background points as well. In the variant we use, each bounding box is randomly shifted and rotated to emulate real-world detection boxes. This process is repeated five times with different bounding box perturbations for each object. This process results in \texttildelow$15$k~objects in total. An object is represented as a cloud of $2048$ 3D spatial points. Additionally, a binary pointwise mask is provided for training, indicating whether a point belongs to the background or to the object. We use the hardest variant of the dataset (PB\_T50\_RS), with the highest rate of bounding box perturbations. We adopt the original~\cite{Uy19} train/test split.

We provide results in Table~\ref{tab:scanobjectnn} and show that our method outperforms the existing ones both in overall and mean-class accuracy. Note, our method outperforms the current state-of-the-art in terms of the overall accuracy by a large margin (\textbf{+3.7\%}). We also evaluate a CT variant with a learnable anisotropic scaling $s$ applied after the transformation $T$ in the \eqref{eq:pred_keys}. The resulting model outperforms the state-of-the-art  by $\textbf{3.5}\%$ in the mean class accuracy.

For completeness, we have also evaluated our approach on the ModelNet40 benchmark~\cite{Wu2015}, which has been saturated. Following the PointNet++ protocol~\cite{Qi2017}, our approach (CT+scales) achieves $93.1$ in Overall Accuracy (OA), and $90.8$ in mean Class Accuracy (mAcc), which is on par or better than other methods that work with point clouds (rather than CAD meshes). 

\begin{table}
\begin{tabular}{|c|c|c|}
\hline
& overall acc. & mean class acc.  \\ \hline
3DmFV \cite{Ben2018}      & 63           & 58.1   \\ \
PointNet \cite{Qi2016PointNetDL}   & 68.2         & 63.4     \\ 
SpiderCNN \cite{Xu2018}  & 73.7         & 69.8  \\ 
PointNet++ \cite{Qi2017} & 77.9         & 75.4    \\ 
DGCNN \cite{Wang19}      & 78.1         & 73.6  \\
PointCNN \cite{pointcnn18}   & 78.5         & 75.1    \\ 
DRNet \cite{Qiu2021a} & 80.3 & - \\ 
GFNet  \cite{Qiu2021}    & 80.5        & 77.8 \\ 
DI-PointCNN \cite{Zhai2020} & 81.3 & 79.6  \\ \hline
\textbf{CT (ours)}  & 85.0  & 80.7 \\ \hline
\textbf{CT (ours) + scales} & \textbf{85.5} & \textbf{83.1} \\ \hline
\end{tabular}
\caption{Classification results on ScanObjectNN. Our method outperforms the others both in terms of overall accuracy and mean class accuracy.\vspace{-3ex}\label{tab:scanobjectnn}}
\end{table}


\paragraph{Single-View Object Reconstruction.} In our generation experiments we follow the recently introduced benchmark \cite{Tatarchenko19} on 3D object reconstruction. The benchmark is based on ShapeNet~\cite{shapenet2015} renderings. Unlike previous ShapeNet-based benchmarks for image-based reconstruction that used \textit{canonical coordinate frames}, the new one argues that the reconstructions should be evaluated in the \textit{viewer-based coordinate frame}, where the task is more challenging and more realistic. The work~\cite{Tatarchenko19} also provides evaluations of several recent methods on image-based reconstruction, as well as the retrieval-based oracle. The dataset consists of ShapeNet \cite{shapenet2015} models, where each model belongs to one of $55$~classes. Each object has been rendered with ShapeNet-Viewer from five random view points. We employ the same train/val/test split as in~\cite{Tatarchenko19}.  

In the benchmark, objects were rendered to $224 \times 224$ pixel images, which we resize to $128 \times 128$ pixels and then fed to our model. The ground truth is represented as a cloud of $10.000$~points in the viewer-aligned coordinate system.

Our model outputs $8196$~points to represent a reconstructed object. Since the protocol requires to predict exactly $10.000$~points, we perform the reconstruction twice with different sphere noise and the same style vector $\vz$ extracted by the convolutional encoder (\fig{generator}). This results in $16.392$~points total from which we randomly select $10.000$~points.

The main evaluation metric proposed in~\cite{Tatarchenko19} is the $F$-score computed at a $1 \%$ volume distance threshold. The methods are compared with macro-averaged $F$-score~@$1\%$ and by the number of classes, in which a method has the highest mean $F$-score~@$1\%$. Our quantitative results are summarized in \tab{reconstruction}. In is evident that our method outperforms all methods evaluated in~\cite{Tatarchenko19} \textit{including} the retrieval-based oracle very significantly.





\begin{table}[!t]
\setlength\tabcolsep{0.5pt}
\begin{center}
\begin{tabular}{|l|c|c|}
\hline
Methods & mAvg. $F$-score@$1\%$\,\,\,\,\, & Top-1 cat.\\
\hline
AtlasNet \cite{groueix2018}  & $0.252$  & $2$ \\
Matryoshka \cite{Richter2018} & $0.217$  & $0$  \\
OGN \cite{Tatarchenko2017} & $0.264$ & 1 \\
Retrieval &  $0.236$ &  $0$ \\  
Retrieval (oracle)  & $0.290$ & $3$ \\
\hline
\textbf{CT (ours)} & $\mathbf{0.359}$ & $\mathbf{49}$ \\\hline
\end{tabular}
\end{center}
\vspace{-3ex}
\caption{F-score evaluation (@1\%) of 3D shape reconstruction in the viewer-based coordinate frame, averaged by categories. The cloud transformer outperforms other methods including the retrieval based oracle considerably. 
}
\label{tab:reconstruction}
\vspace{-3ex}
\end{table}

\paragraph{Indoor Semantic Segmentation.} The Stanford Indoor Dataset (S3DIS)~\cite{armeni_cvpr16} is a popular 3D point cloud segmentation benchmark that consists of large 3D point cloud scenes captured at three different buildings annotated with 13 semantic labels at the point level. The dataset comes with six splits. For the sake of fair comparison, we evaluate on S3DIS using a conventional protocol, established by \cite{Qi2016PointNetDL}, which chunks rooms into 1m$\times$1m blocks. Each block consists of $4096$ points and each point is represented with its 3D coordinates and its RGB color, which results in a six-dimensional input vector.
In this setting, the average inference time is 0.5 sec on a Tesla P40 GPU card, with only 200 MB memory per chunk used. Following many previous works, we evaluate on the `Area 5' split and train on the remaining five splits, as \cite{Tchapmi2017} advocate this fold as representative in measuring generalization ability due to being shot in a separate building. 

Since the current published state-of-the-art-method JSENet~\cite{Hu2020}, uses a different protocol~\cite{Thomas2019KPConvFA}), we also evaluate our model using 'KPConv' protocol. In it, at each step an input point cloud is dynamically sampled from a sphere of radius 2$m$. Each point cloud contains up to $8192$ points. During evaluation, the same data strategy is applied together with voting.

In the standard 1m $\times$ 1m protocol $xyz$ spatial coordinates are augmented with random rotation, anisotropic scale, jitter and shifts. For color, on thhe other hand, we use chromatic autocontrast, jitter and translation (following \cite{Choy20194DSC}). As for the 'KPConv' protocol, we follow the original paper and augment point clouds with anisotropic random scaling, spatial gaussian jittering, random rotations around the z-axis and random color dropping. \tab{s3dis} shows that for the semantic segmentation task our method outperforms state-of-the-art in both considered protocols.


\begin{table}

\begin{center}
\begin{tabular}{|l|c||l|c|}
\hline 
Method & mIoU & Method & mIoU\\
\hline
PointNet \cite{Qi2016PointNetDL}	    & $41.1$ &  SPGraph*  \cite{Landrieu2018LargeScalePC}	& $58.0$	\\
Eff. 3D Conv \cite{zhang2018efficient}	& $51.8$ & 	Minkowski32* \cite{Choy20194DSC}            & $65.3$    \\
RNN Fusion \cite{ye20183d}              & $57.3$ & 	KPConv$\dagger$ \cite{Thomas2019KPConvFA}   & $67.1$    \\
ParamConv \cite{WangParam18}          	& $58.3$ & 	JSENet$\dagger$ \cite{Hu2020}               & $67.7$    \\
PointCNN \cite{pointcnn18}              & $57.2$ &  Point Trans.* \cite{Zhao2021}               & $\mathbf{70.4}$  \\
\hline
\textbf{CT (ours)} & $63.7$ & \textbf{CT$\dagger$ (ours)}  &  $67.9$   \\
\hline
\end{tabular}
\end{center}
  \vspace{-2.5ex}
\caption{Semantic segmentation intersection-over-union scores on S3DIS \textit{Area-5} split. The methods in the left column use the standard protocol with chunking of the scene into blocks, methods marked with $\dagger$ employ KPConv's~\cite{Thomas2019KPConvFA} protocol. Other protocols are labeled with $*$. Cloud transformers outperform state-of-the-art in standard protocol and perform better than previously published works in other protocols (Note that Point Transformer is a concurrent line of work).}
\label{tab:s3dis} 
\vspace{-2.5ex}
\end{table}

\vspace{-1ex}
\paragraph{Point Cloud Completion.} Finally, we evaluate Cloud Transformer Inpainter on the ShapeNet-based benchmark~\cite{yuan2018} for high-resolution point cloud completion. The benchmark is composed of the eight largest ShapeNet categories (airplane, cabinet, car, chair, lamp, sofa, table, vessel). This makes $30974$ unique objects in total. In each category $100$ of unique objects is reserved for validation and $150$ for testing. The dataset consists of $(P_{\text{part}}, P_{\text{gt}})$ pairs, where $P_{\text{part}}$ is a \textit{partial} point cloud and $P_{\text{gt}}$ is a \textit{complete} point cloud. Partial point clouds are obtained via 2.5D depth image back-projection, taken from a random view. There are eight random views generated per each training object. The partial cloud consists of no more than $2048$~points, while the complete point cloud consists of $16.384$~points. In contrast to the image-based reconstruction, both partial and complete point clouds are provided in the object-based coordinate system.

\begin{table}
  \begin{center}
  \begin{tabular}{|l|c|c|}
    \hline
    Methods     & mAvg. F-Score@1\% &  mAvg. CD \\
    \hline  
    AtlasNet \cite{groueix2018}    
                 & 0.616 & 4.523  \\
    PCN \cite{yuan2018}
                 & 0.695 & 4.016 \\
    FoldingNet \cite{Yang2018}
                 & 0.322 & 7.142 \\
    TopNet \cite{Tchapmi2019}
                 & 0.503 & 5.154 \\
    MSN \cite{Liu2019m}
                 & 0.705 & 4.758 \\
    GRNet  \cite{Xie2020}     & 0.708 & \textbf{2.723}  \\ \hline
    \textbf{CT (ours)}  & \textbf{0.752} & 3.392  \\
  	\hline
  \end{tabular}
  \vspace{-3ex}
  \end{center}
\caption{Point completion results on ShapeNet compared using F-Score@1\% and CD. Note that both of them are computed on 16,384 points and macro-averaged.  \label{tab:completion}}  
\end{table}

We predict a high-resolution reconstruction with $16.384$ points in a single pass of our Cloud Transformer Inpainter network. Our model produces detailed reconstructions with complex geometries (see Figure~\ref{fig:teaser}). Regarding quantitative evaluation of our method, we report both F-Score@1\% and CD (Chamfer Distance), both of them computed with the ground truth $16.384$ point clouds. Following~\cite{Tatarchenko19}, we argue that the F-Score@1\% should be regarded as a primal metric of the shape prediction quality, and in this metric our method again beats state-of-the-art considerably (\tab{completion}). 


\paragraph{Ablation Study.} We also perform an ablation study to justify our architecture choices. We consider the following ablations:
\begin{itemize}
    \setlength\itemsep{0.01em}
    \item \textbf{Linear key prediction}: We replace key prediction procedure with a linear point-wise layer $d$, followed by BatchNormalization. Using the notations form \ref{sect:method}, $\vk_i = d(\vx_i)$.
    \item \textbf{Mean agg.} and \textbf{Sum agg.}: The aggregation method in the rasterization step is replaced with element-wise mean and sum correspondingly. Note, in the latter case (sum) it makes our operation similar to the splatting, used in SplatNet~\cite{Su18}. 
    \item \textbf{Non-learnable keys:} In this ablation we use different non-learnable projections of the input positions as keys. More precisely,  $\vk_i = T \left(\vp_i \right)$, where $T \in SO(3)$ is a \textit{fixed} random transformation and no deep residual predicted. While this variant performs on par with \textit{linear key prediction} for segmentation, it performs marginally better on classification.
    \item We also train an architecture \textbf{without planar heads} to see if using only volumetric heads might be sufficient.
    \item In the \textbf{Coarser feature maps} experiment the spatial dimensions of feature maps are halved.
    \item \textbf{No multihead}: We ablate our multi-headed architecture by replacing 16 headed CMHCT blocks with a single-headed block, where we increase the channel dimension $8$~times to keep the model's capacity intact. 
    \item Finally, we consider  \textbf{2x shallower} and architectures, replacing four CMHCT blocks with two CMHCT blocks.
\end{itemize}
\begin{table}[h]
\vspace{-3ex}
\begin{center}
\begin{tabular}{|l|c|c|}
\hline
Methods & mIOU S3DIS & acc. ScanObjNN \\
\hline
Linear key prediction & 62.5 & 81.9  \\ 
Sum aggregation & 57.9 & 82.1 \\
Mean aggregation & 61.3 & 83.4 \\ 
Without planar heads  & 63.4 & 84.8\\
Coarser feature maps & 62.2 & 84.4 \\ 
No multihead & 63.1 & 84.0 \\ 
Non-learnable keys & 62.5 & 84.9 \\
2x shallower & 62.2 & 84.1 \\  
\hline
\textbf{CT (full)} & 63.7 & 85.0 \\
\hline
\end{tabular}
\end{center}
\vspace{-3ex}
\caption{Ablation study on S3DIS semantic segmentation and ScanObjectNN classification. See text for discussion.}
\label{tab:ablation}
\vspace{-3ex}
\end{table}

Observing the advantage of the full architecture over the shallower architecture in the S3DIS case, we have also evaluated a 2x deeper architecture with eight CMHCT blocks, achieving 64.1 mIOU score. With (T = Id) (see \ref{fig:cloud_transform}) ablation we observed $61.9$ mIOU thus learnable projections are of great importance. Our preliminary ablation with $d=0$ suggests that the effect is small for segmentation (drop of 0.04\%), but we expect it to be higher for generative tasks where the input point cloud is trivial (spherical). 

We have also evaluated the importance of the \textbf{gradient balancing trick}. On the S3DIS an ablation without gradient balancing trick achieved 62.8 mIOU. More importantly, when we tried to run the linear key prediction variant without the gradient balancing, the learning diverged for all reasonable learning rates, revealing the importance of gradient balancing.

\section{Conclusion}
\label{sect:conclusion}
We have presented a new block for neural architectures that process point clouds. Our block extends the ideas of Spatial Transformers, Transformers, and Multi-View CNNs on neural point cloud processing. 

While there are some significant differences between our architecture and the Transformer, we want to highlight some interesting similarities. Both Transformer and our architecture operate on sets and use parallel heads. Most importantly, similarly to Transformer, our architecture achieves quick and long range information propagation without blowing up the number of learnable parameters. In the semantic segmentation $1\times1$ chunk case, an average point ``interacts'' with 39\% of other points after the first MHCT block and with 100\% of points after just one CMHCT block (i.e.~just three MHCT blocks).

Based on the new block, we have presented architectures for point cloud semantic segmentation, point cloud classification, point cloud completion and single-image based geometry reconstruction that achieve state-of-the-art results. 



\newpage
\appendix

\twocolumn[{%
\renewcommand\twocolumn[1][]{#1}%
    \centering
    \includegraphics[width= \textwidth]{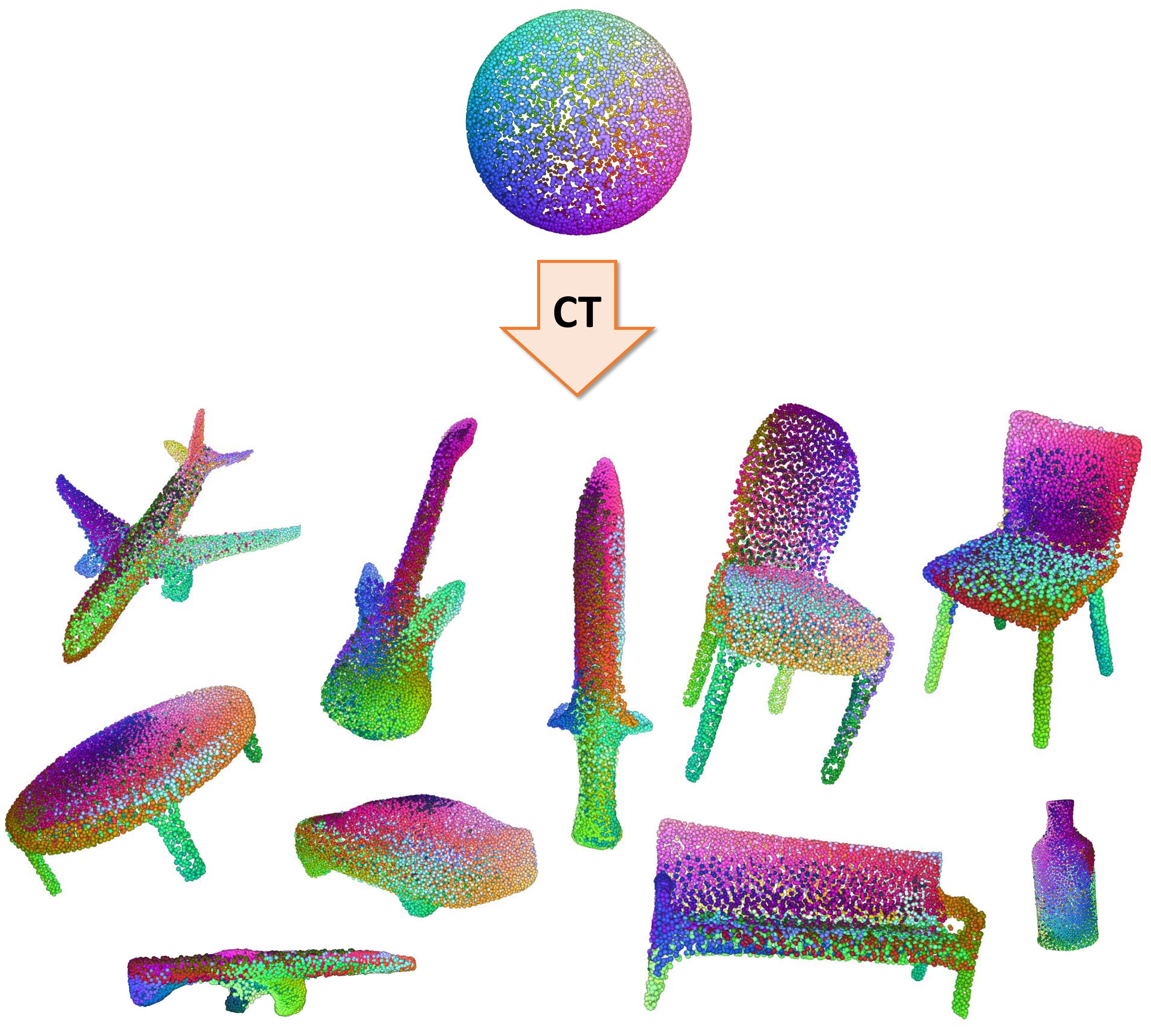}
    \captionof{figure}{Results of image-based reconstruction. Our network performs the reconstruction by sampling points from the unit sphere $S^{2}$ (top) and then ``folding'' this set into a point cloud corresponding to the input image. Here, we show the point clouds, with points colored according to their initial positions on the sphere (red, green, and blue values are used to color-code each Cartesian coordinate). The color-coding reveals shape correspondences within and across classes.}
    \label{fig:latent_viz}
    \vspace{2ex}
}]


\section{Cloud Transform pseudocode}

We provide a pseudo-code in Algorithm~\ref{alg:ct} as well as the PyTorch code: \href{https://saic-violet.github.io/cloud-transformers/}{https://saic-violet.github.io/cloud-transformers/}. 

\begin{algorithm}
 \SetKwFunction{Convolve}{Convolve}
  \SetKwFunction{Linear}{Linear}

 \KwData{Input point cloud positions $P \in \RR^{3 \times n}$, their high-dimensional input features $X \in \RR^{f \times n}$}
 \KwResult{New high-dimensional output features $Y \in \RR^{g \times n}$}
 Initialize a feature-map $I \in \RR^{w \times w \times c_{in}}$ of spatial dimension $w$ with zeros;\

 \For{each point $\vp_i \in P$}{

  Predict residuals $\vr_i \leftarrow \Linear(\vp_i)$\;
  
  Compute rasterization positions $\vk_i \leftarrow T(\vp_i + \vr_i)$ (see eq. (1) of the main paper);\
  
  Predict feature to rasterize $\vv_i \leftarrow \Linear(\vx_i)$\;
  
  Compute bilinear coordinates $b_i$ and rasterization positions $[h_{\bullet}, w_{\bullet}]$ of $\vk_i$ wrt $I$ (see eq. (1) of the main paper)\;
  
  \BlankLine
  
  \tcp{rasterization}
  $I[h_0, w_0]  \leftarrow  \operatorname{max}(I[h_0, w_0], b_i^{00} \vv_i) $ \; 
  $I[h_0, w_1]  \leftarrow  \operatorname{max}(I[h_0, w_1], b_i^{01} \vv_i) $ \;
  $I[h_1, w_1]  \leftarrow  \operatorname{max}(I[h_1, w_0], b_i^{10} \vv_i) $ \;
  $I[h_1, w_1]  \leftarrow  \operatorname{max}(I[h_1, w_1], b_i^{11} \vv_i) $ \;
 }
 
 $I \leftarrow $ \Convolve ($I$)\;

 \For{each point $\vp_i \in P$}{
 
  \tcp{de-rasterization}

  $\tilde{\vv_i} \leftarrow b_i^{00} I[h_0, w_0] +  b_i^{01} I[h_0, w_1] + b_i^{10} I[h_1, w_1] +  I[h_1, w_1] b_i^{11}$ ;\
  
   Predict output features $\vy_i \leftarrow \Linear(\tilde{\vv_i})$ \;
 }
 \Return{$Y = \{ \vy_0, \ldots, \vy_{n - 1} \} $}
 
 \caption{Cloud Transform pseudo-code. For the full discussion please refer to the main manuscript, Section 3.1}
 \label{alg:ct}
\end{algorithm}

\section{Additional experimental results}

In \fig{latent_viz}, we show the
 ``foldings'' of the input sphere in our Single-View reconstruction experiments. The resulting point colors represent spatial coordinates of their input positions on the sphere. The shape correspondences learned implicitly during training are revealed.
Additionally, we provide per-class results for our experiments in \tab{s3dis-cl}, \tab{scanobjectnn-cl},  \tab{shapenet-completion-fscore-cl}, \tab{shapenet-completion-cd-cl} and \tab{im2p_perclass_results}.

\newcolumntype{L}[1]{>{\raggedright\let\newline\\\arraybackslash\hspace{0pt}}m{#1}}
\newcolumntype{C}[1]{>{\centering\let\newline\\\arraybackslash\hspace{0pt}}m{#1}}
\newcolumntype{R}[1]{>{\raggedleft\let\newline\\\arraybackslash\hspace{0pt}}m{#1}}

\section{Experimental details} 

We train our models using the standard ADAM optimizer~\cite{kingma15} with the learning rate $1e{-}4$ for generative tasks (image-based reconstruction and point cloud completion) and with the learning rate $1e{-}3$ for recognition tasks (semantic segmentation and classification). We halve the learning rate every $100$k iterations for image-based reconstruction and point cloud inpainting experiments. For classification and semantic segmentation experiments, the learning rate is decayed by $0.7$ and every $25$k iterations.

For classification and segmentation experiments we use batch size $8$ per GPU and $2$ GPUs in total. For image-based reconstruction and point cloud completion, the batch size $4$ and $2$ are used respectively. In both cases, we train the models on eight GPUs.

\section{Multi-Headed Cloud Pooling}

\begin{table}
\begin{tabular}{|c | c|} 
\hline
\textbf{3D CNN} & \textbf{2D CNN} \\ [0.5ex]
\hline 
Res3D(in=32, out=64) & Res2D(in=16, out=32) \\ 
\hline
MaxPool(2) & MaxPool(2) \\
\hline
Res3D(in=64, out=64) & Res2D(in=32, out=64)   \\ 
\hline
MaxPool(2) & MaxPool(2)  \\ 
\hline
Res3D(in=64, out=64) & Res2D(in=64, out=64)  \\ 
\hline
AvgPool & AvgPool  \\ 
\hline
\end{tabular}
\caption{Our mini-CNN architectures, in the 2D and 3D cases. Note that a different ConvNet applied per each head independently.}
\label{tab:cnn_arch}
\end{table}

\begin{figure*}[h]
\includegraphics[width=\textwidth]{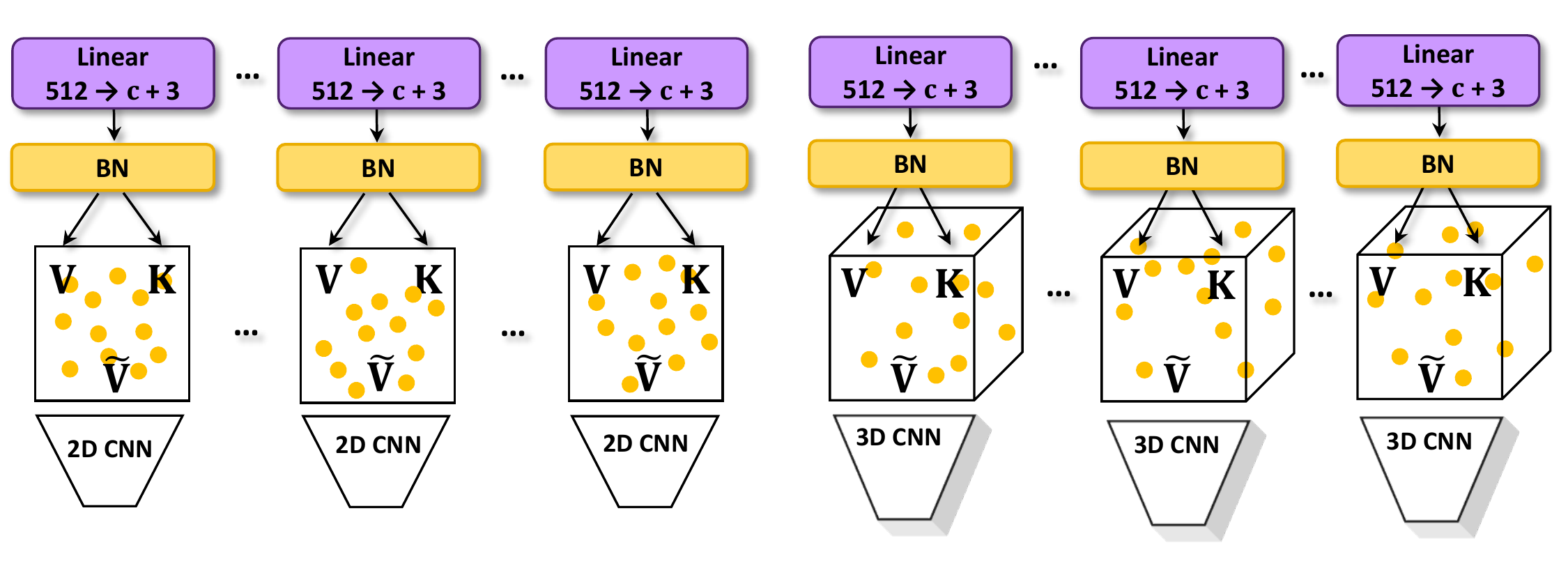}
\caption{The Multi-Headed Cloud Pooling layer operates as a Multi-Headed Cloud Transform layer except for the de-rasterization step. Instead of de-rasterization, a compact 2D or 3D ConvNet is applied producing a single vector as an output of each head.}
\label{fig:cloud_pool}
\end{figure*}

 Here we provide more details on our multi-headed cloud pooling layer used in our classification model.
 
To solve the classification task, we introduce a multi-headed pooling layer (\fig{cloud_pool}). This is needed, because unlike other tasks, the output of the classification process is a global vector of probabilities. We devise this layer to be as similar to our standard layers as possible.

Thus, similarly to the regular MHCT layer, the new layer performs multiple rasterizations onto 2D and 3D feature maps of spatial size $8$ and $16$ respectively. The channel dimension of each head is $32$ for three-dimensional heads and $16$ for two-dimensional heads. Afterward, the resulting feature maps are processed with three standard convolutional residual blocks, each interchanged with a max-pooling.  The architectures of these 2D and 3D ConvNets are shown in \tab{cnn_arch}. The residual path of Res(in, out) consists of two convolutional layers Conv(in, out) and Conv(out, out) with a $3 \times 3$ (or $3 \times 3 \times 3$) spatial filters, each followed by a BatchNormalization and ReLU activation.   

The resulting vectors are aggregated across the heads via concatenation and processed with a dense layer to form a final classification vector $\vk_{class}$.

\section{Gradient Balancing}
Below we provide a more thorough discussion of our gradient balancing trick introduced in the main paper and its necessity. 

\paragraph{Problem discussion}
Ultimately, the problem can be pinpointed to the fact that as the key $\vk_i$ moves from the top-left to the bottom-right of a certain grid cell thus traversing only $1/w$-th of its variation range, the assignment weight of $\vv_i$ to the bottom-right corner changes from $0$ to $1$ (i.e\ traverses the full variation range). This means that gradients w.r.t.\ keys $\vk_i$ in our architecture will always be roughly $w$ times stronger than those w.r.t.\ values $\vv_i$.

We justify this trick by an exact derivation. 
Further we denote a target loss function as $\mathcal{L}$ and assume the spatial feature dimension to be $w{-}1$.

\begin{lemma}
\label{lemma:grad_form}
Let $\vk = (k_0, k_1) \in [0, 1]^2$ be a key, which is typically an output of the network's key prediction branch in our Cloud Transform block. Let $\vb$ be a vector of bilinear weights of $\vk$  inside the enclosing cell, as in equation (1) (main paper). Then: 

\begin{equation}
\setlength\arraycolsep{2pt}
\small
    \dpart{\vb}{\vk}  = w \cdot \begin{pmatrix}
        (w \cdot k_1{-}\lceil w \cdot k_1 \rceil ) 
        & &  
        (w \cdot k_0{-}\lceil w \cdot k_0 \rceil)   \\
        
     {-} (w \cdot k_1{-}\lfloor w \cdot k_1 \rfloor ) 
        & &
       {-} (w \cdot k_0{-}\lceil w \cdot k_0 \rceil )   \\ 
     
       {-} (w \cdot k_1{-}\lceil w \cdot k_1 \rceil)
        & & 
        {-}  (w \cdot k_0{-}\lfloor w \cdot k_0 \rfloor )  \\
        
         (w \cdot k_1{-}\lfloor w \cdot k_1 \rfloor ) 
         & &  
        (w \cdot k_0{-}\lfloor w \cdot k_0 \rfloor )   \\ 
\end{pmatrix}
\end{equation}
\end{lemma}

The derivation is straightforward, given the formula  for the bilinear weights $\vb=\left( b^{00}, b^{01}, b^{10}, b^{11} \right)$: 
\begin{align}
\begin{split}
\label{eq:bill}
    b^{00} &=  \; \, (w {\cdot} k_i^0 - h_1)(w{\cdot}k_i^1 - w_1)  \\
    b^{01} &= - (w{\cdot}k_i^0 - h_1)(w{\cdot}k_i^1 - w_0)  \\
    b^{10} &= - (w{\cdot}k_i^0 - h_0)(w{\cdot}k_i^1 - w_1)  \\
    b^{11} &=   \; \, (w{\cdot}k_i^0 - h_0)(w{\cdot}k_i^1 - w_0) 
\end{split}
\end{align}

To ease the notations, we denote the matrix above as $D$, i.e $\dpart{\vb}{\vk}  = w \cdot D$.

\newcommand{\Varr}[1]{\operatorname{Var}\left( {#1} \right)}
\newcommand{\Trace}[1]{\operatorname{tr}\left( {#1} \right)}

\begin{lemma}
\label{lemma:grad_norm}
Let $X =  \dpart{\mathcal{L}}{\vb_i}$ and \, $\Varr{X}$ be its variance. Then the following inequality on matrix spectral norms holds:

\begin{equation}
  ||D \Varr{X} D^T ||_2 \geq \frac{1}{2} ||\Varr{X}||_2
\end{equation}


\begin{proof}
\begin{equation}\label{my eq}
\begin{aligned}
  & 2 {||D \Varr{X} D^T ||}_2 && \text{} \\
  & \geq \Trace{D \Varr{X} D^T }  && \text{trace cyclic prop.} \\
  & \geq \Trace{D D^T \Varr{X}} && \text{} \\
  & \geq ||D D^T \Varr{X}||_2  && \text{}
  \\
  & \geq  \sigma_n(D D^T) ||\Varr{X}||_2 && \\
 \\
\end{aligned}
\end{equation}
, where $\sigma_n(D D^T)$ is the smallest non-zero singular value of $D^T D$.

Note that $D$ has a special form, given $-1 \leq a, b \leq 0$:
\begin{equation}
\setlength\arraycolsep{2pt}
\small
    D  = \begin{pmatrix}
        a
        & &  
      b  \\
        
     {-} (1 + a) 
        & &
      {-} b  \\ 
     
      {-} a
        & & 
        {-} (1 + b)  \\
         1 + a 
         & &  
      1 + b  \\ 
\end{pmatrix}
\end{equation}


By a straightforward derivation, one can find two non-zero singular values $\sigma_2 = 1$ and $\sigma_1 = (2a + 1)^2 + (2b + 1)^2 + 1$ of $D D^T$. And $\sigma_1 > \sigma_1$, therefore:

\begin{equation}
  ||D \Varr{X} D^T ||_2 \geq \frac{1}{2} ||\Varr{X}||_2
\end{equation}

This, informally, means that $D$ does not decrease gradient variance $\Varr{X}$ too much. In the case of 3D rasterization operation, the computation is analogous. 
\end{proof}
\end{lemma}

To sum up the results above, the back-propagation from bilinear weights $\vb_i$ to $\vk_i$ has a form:

\begin{equation}
    \dpart{\mathcal{L}}{\vk_i} =
    \left( \dpart{\vb_i}{\vk_i} \right)^T
    \cdot \dpart{\mathcal{L}}{\vb_i} =
    w \cdot D^T \cdot \dpart{\mathcal{L}}{\vb_i}
\end{equation}

From the Lemma~\ref{lemma:grad_norm} we know that multiplication by $D$ does not decrease variance's $\Varr{X}$ spread more than by two times, while $w$ takes typical vales between $16$ and $128$.

Intuitively, the gradients' variance is scaled up by $w$ at each layer during the backpropagation through the keys. Therefore, given a network with $d$ cloud transform layers, the gradient variance would ``explode'' as $w^d$. We have observed such explosions experimentally. The balancing trick discussed above successfully fixes this problem and allows training deep architectures built out of Cloud Transformer blocks.

\begin{table*}
\begin{small}
\begin{center}
\scalebox{0.9}{
\begin{tabular}{|l|ccccccccccccc|}
\hline
Method  & ceil. & floor  & wall & beam & col. & wind. & door & chair & table & book.  & sofa & board & clut. \\
\hline
Pointnet  	& $88.8$	& $97.3$	& $69.8$	& $0.1$	& $3.9$	& $46.3$	& $10.8$	& $52.6$	& $58.9$	& $40.3$	& $5.9$	& $26.4$	& $33.2$ \\
SegCloud* 	& $90.1$	& $96.1$	& $69.9$	& $0.0$	& $18.4$	& $38.4$	& $23.1$	& $75.9$	& $70.4$	& $58.4$	& $40.9$	& $13.0$	& $41.6$	\\
Eff. 3D Conv	& $79.8$	& $93.9$	& $69.0$	& $0.2$	& $28.3$	& $38.5$	& $48.3$	& $71.1$	& $73.6$	& $48.7$	& $59.2$	& $29.3$	& $33.1$	\\
TangentConv 	& $90.5$	& $97.7$	& $74.0$	& $0.0$	& $20.7$	& $39.0$	& $31.3$	& $69.4$	& $77.5$	& $38.5$	& $57.3$	& $48.8$	& $39.8$	\\
RNN Fusion	& $92.3$ & $98.2$	& $79.4$	& $0.0$	& $17.6$	& $22.8$	& $62.1$	& $74.4$	& $80.6$	& $31.7$	& $66.7$	& $62.1$	& $56.7$	\\
SPGraph* 		& $89.4$	& $96.9$	& $78.1$	& $0.0$	& $42.8$	& $48.9$	& $61.6$	& $84.7$	& $75.4$	& $69.8$	& $52.6$	& $2.1$	& $52.2$	\\
ParamConv & $92.3$	& $96.2$	& $75.9$	& $0.3$	& $6.0$	& $69.5$	& $63.5$	& $66.9$	& $65.6$	& $47.3$	& $68.9$	& $59.1$	& $46.2$	\\
PointCNN  &  $92.3$ & $98.2$ & $79.4$ & $0.0$ & $17.6$ & $22.7$ &  $62.1$ &  $74.4$ &  $80.6$ &  $31.7$ &  $66.7$ &  $62.1$ &  $56.7$ \\
\hline
\textbf{CT (ours) std. prot.}  & $94.4$ & $98.2$ & $85.4$ & $0.0$ & $23.6$ & $47.4$ & $71.0$ & $87.3$ & $77.5$ & $66.0$ & $49.3$ & $71.1$ & $57.8$ \\
\hline
Minkowski32*  & $91.7$ & $98.7$ & $86.1$ & $0.0$ & $34.0$ & $48.9$ & $62.4$ & $89.8$ &  $81.5$ & $74.8$ & $47.2$ & $74.4$ & $58.5$ \\
\hline
KPConv$\dagger$  & $92.8$ & $97.3$ & $82.4$	& $0.0$	& $23.9$ & $58.0$ & $69.0$ & $91.0$ & $81.5$	& $75.3$ & $75.4$	& $66.7$ & $58.9$ \\
JSENet$\dagger$ & $93.8$ & $97.0$ & $83.0$ & 0.0 & $23.2$ & $61.3$ & $71.6$ & $89.9$ & $79.8$ & $75.6$ & $72.3$ & $72.7$ & $60.4$ \\ \hline
\textbf{CT$\dagger$ (ours)} & $94.2$ & $97.7$ & $82.7$  & $0.0$ & $34.4$ & $62.8$ & $68.4$ & $89.8$ & $80.4$ & $78.2$ & $61.4$ & $67.7$ & $64.9$ \\  \hline
\end{tabular}}
\end{center}
\end{small}
\caption{Semantic segmentation intersection-over-union scores on S3DIS \textit{Area-5} split. The models without any marks use the standard protocol with chunking of the scene into blocks, while the models with $\dagger$ employ KPConv's~\cite{Thomas2019KPConvFA} protocol. The rest protocols are labeled with $*$. Per-class results are provided.}
\label{tab:s3dis-cl} 
\end{table*}

\begin{table*}
\begin{center}
\caption{Per-class classification results on ScanObjectNN.}
\resizebox{\textwidth}{!}{
\begin{tabular}{|c|c c c c c c c c c c c c c c c|}
\hline
 & bag  & bin  & box  & cabinet & chair & desk & display & door & shelf & table & bed  & pillow & sink & sofa & toilet \\ \hline
3DmFV              & 39.8 & 62.8 & 15.0 & 65.1    & 84.4  & 36.0 & 62.3    & 85.2 & 60.6  & 66.7  & 51.8 & 61.9   & 46.7 & 72.4 & 61.2   \\ \hline
PointNet          & 36.1 & 69.8 & 10.5 & 62.6    & 89.0  & 50.0 & 73.0    & 93.8 & 72.6  & 67.8  & 61.8 & 67.6   & 64.2 & 76.7 & 55.3   \\ \hline
SpiderCNN            & 43.4 & 75.9 & 12.8 & 74.2    & 89.0  & 65.3 & 74.5    & 91.4 & 78.0  & 65.9  & 69.1 & 80.0   & 65.8 & 90.5 & 70.6   \\ \hline
PointNet++           & 49.4 & 84.4 & 31.6 & 77.4    & 91.3  & 74.0 & 79.4    & 85.2 & 72.6  & 72.6  & 75.5 & 81.0   & 80.8 & 90.5 & 85.9   \\ \hline
DGCNN              & 49.4 & 82.4 & 33.1 & 83.9    & 91.8  & 63.3 & 77.0    & 89.0 & 79.3  & 77.4  & 64.5 & 77.1   & 75.0 & 91.4 & 69.4   \\ \hline
PointCNN            & 57.8 & 82.9 & 33.1 & 83.6    & 92.6  & 65.3 & 78.4    & 84.8 & 84.2  & 67.4  & 80.0 & 80.0   & 72.5 & 91.9 & 71.8   \\ \hline

GFNet   & 59.0 & 84.4 & 44.4 & 78.2    & 92.1  & 66 & 91.2   & 91.0 & 86.7  & 70.4  & 82.7 & 78.1   & 72.5 & 92.4 & 77.6   \\ \hline

DI-PointCNN  & 65.1 & 80.9 & 62.4 & 80.4 & 90.5 & 78.0 & 86.8 & 88.1 & 84.6 & 67.0 & 84.5 & 82.8 & 73.3 & 95.2 & 74.1  \\ \hline

\textbf{CT (ours)}   
& 50.6 &  86.9 & 52.6 & 87.4 & 96.2 &  77.3 &
 85.8 & 93.8 & 88.4 &  80.0  &  76.4 & 87.6 & 81.7 & 94.3 & 89.4  \\ \hline
\textbf{CT (ours) + scales}   & 57.8 & 89.9 & 45.9 & 87.4 & 95.6 & 84.7 & 
 84.8 & 92.9 & 88.4 & 77.4 & 82.7 & 85.7 &  85.8 & 92.9 & 95.3 \\
\hline
\end{tabular}
\label{tab:scanobjectnn-cl}
}
\end{center}
\end{table*}

\begin{table*}[!t]
  \begin{center}
  \caption{Per-class point completion results on ShapeNet compared using F-Score@1\%. Note that the F-Score@1\% is computed on 16,384 points.}
  \begin{tabular}{|l|cccccccc|c|}
    \hline
    Methods      & Airplane   & Cabinet    & Car        & Chair 
                 & Lamp       & Sofa       & Table      & Watercraft \\
    \hline  
    AtlasNet     
                 & 0.845      & 0.552      & 0.630      & 0.552
                 & 0.565      & 0.500      & 0.660      & 0.624 \\
    PCN 
                 & 0.881      & 0.651      & 0.725 & 0.625
                 & 0.638      & 0.581      & 0.765      & 0.697 \\
    FoldingNet 
                 & 0.642      & 0.237      & 0.382      & 0.236
                 & 0.219      & 0.197      & 0.361      & 0.299  \\
    TopNet 
                 & 0.771      & 0.404      & 0.544      & 0.413
                 & 0.408      & 0.350      & 0.572      & 0.560 \\
    MSN
                 & 0.885 & 0.644      & 0.665      & 0.657 
                 & 0.699      & 0.604      & 0.782 & 0.708 \\
    GRNet        & 0.843      & 0.618      & 0.682      & 0.673
                 & 0.761      & 0.605      & 0.751      & 0.750 \\ \hline
    \textbf{CT (ours)}    & 0.921 &	0.652 & 0.733 & 0.710 & 0.774 & 0.628 & 0.811 & 0.789 \\
  	\hline
  \end{tabular}
  \label{tab:shapenet-completion-fscore-cl}
  \end{center}
\end{table*}


\begin{table*}[!t]
  \begin{center}
  \caption{Per-class point completion results on ShapeNet compared using Chamfer Distance (CD) with L2 norm computed on $16.384$ points and multiplied by $10^4$. }
  \begin{tabular}{|l|cccccccc|c|}
    \hline
    Methods      & Airplane   & Cabinet    & Car        & Chair 
                 & Lamp       & Sofa       & Table      & Watercraft  \\
    \hline
    AtlasNet  
                 & 1.753      & 5.101      & 3.237      & 5.226
                 & 6.342      & 5.990      & 4.359      & 4.177 \\ 
    PCN 
                 & 1.400 & 4.450      & 2.445 & 4.838
                 & 6.238      & 5.129      & 3.569      & 4.06 \\ 
    FoldingNet 
                 & 3.151      & 7.943      & 4.676      & 9.225
                 & 9.234      & 8.895      & 6.691      & 7.32 \\ 
    TopNet 
                 & 2.152      & 5.623      & 3.513      & 6.346
                 & 7.502      & 6.949      & 4.784      & 4.359 \\ 
    MSN 
                 & 1.543      & 7.249      & 4.711      & 4.539
                 & 6.479      & 5.894      & 3.797      & 3.853 \\ 
    GRNet  & 1.531      & 3.620 & 2.752      & 2.945
                 & 2.649 & 3.613 & 2.552 & 2.122 \\ \hline
    \textbf{CT (ours)} & 1.059 & 4.592 & 2.581 & 4.163 & 3.294 & 5.816 & 3.360 & 2.274   \\
    \hline
  \end{tabular}
  \label{tab:shapenet-completion-cd-cl}
  \end{center}
\end{table*}

\begin{table*}
\small
\begin{center}
\begin{tabular}{|l|c c c c c| c|}
\hline
 & \textbf{AtlasNet} & \textbf{OGN} & \textbf{Matryoshka} & \textbf{Retrieval} & \textbf{Oracle NN} & \textbf{CT(ours)}\\ 
\hline
airplane & 0.39 & 0.26 & 0.33 & 0.37 & 0.45 & 0.53 \\
ashcan & 0.18 & 0.23 & 0.26 & 0.21 & 0.24 & 0.33 \\
bag & 0.16 & 0.14 & 0.18 & 0.13 & 0.15 & 0.20 \\
basket & 0.19 & 0.16 & 0.21 & 0.15 & 0.15 & 0.22 \\
bathtub & 0.25 & 0.13 & 0.26 & 0.22 & 0.26 & 0.33  \\
bed & 0.19 & 0.12 & 0.18 & 0.15 & 0.17 & 0.22 \\
bench & 0.34 & 0.09 & 0.32 & 0.3 & 0.34 & 0.46 \\
birdhouse & 0.17 & 0.13 & 0.18 & 0.15 & 0.15 & 0.31 \\
bookshelf & 0.24 & 0.18 & 0.25 & 0.2 & 0.2 & 0.29 \\
bottle & 0.34 & 0.54 & 0.45 & 0.46 & 0.55 & 0.59 \\
bowl & 0.22 & 0.18 & 0.24 & 0.2 & 0.25  & 0.25  \\
bus & 0.35 & 0.38 & 0.41 & 0.36 & 0.44 & 0.53 \\
cabinet & 0.25 & 0.29 & 0.33 & 0.23 & 0.27 & 0.44 \\
camera & 0.13 & 0.08 & 0.12 & 0.11 & 0.12 & 0.18 \\
can & 0.23 & 0.46 & 0.44 & 0.36 & 0.44 & 0.48 \\
cap & 0.18 & 0.02 & 0.15 & 0.19 & 0.25 & 0.16 \\
car & 0.3 & 0.37 & 0.38 & 0.33 & 0.39 & 0.45 \\
cellular & 0.34 & 0.45 & 0.47 & 0.41 & 0.5 & 0.58 \\
chair & 0.25 & 0.15 & 0.27 & 0.2 & 0.23 & 0.35 \\
clock & 0.24 & 0.21 & 0.25 & 0.22 & 0.27 & 0.36 \\
dishwasher & 0.2 & 0.29 & 0.31 & 0.22 & 0.26 & 0.27 \\
display & 0.22 & 0.15 & 0.23 & 0.19 & 0.24 & 0.31 \\
earphone & 0.14 & 0.07 & 0.11 & 0.11 & 0.13 & 0.27 \\
faucet & 0.19 & 0.06 & 0.13 & 0.14 & 0.2 & 0.30 \\
file & 0.22 & 0.33 & 0.36 & 0.24 & 0.25 & 0.43  \\
guitar & 0.45 & 0.35 & 0.36 & 0.41 & 0.58 & 0.60 \\
helmet & 0.1 & 0.06 & 0.09 & 0.08 & 0.12 & 0.13 \\
jar & 0.21 & 0.22 & 0.25 & 0.19 & 0.22 & 0.28 \\
keyboard & 0.36 & 0.25 & 0.37 & 0.35 & 0.49 & 0.32  \\
knife & 0.46 & 0.26 & 0.21 & 0.37 & 0.54 & 0.61 \\
lamp & 0.26 & 0.13 & 0.2 & 0.21 & 0.27 & 0.37 \\
laptop & 0.29 & 0.21 & 0.33 & 0.26 & 0.33 & 0.44 \\
loudspeaker & 0.2 & 0.26 & 0.27 & 0.19 & 0.23 & 0.33 \\
mailbox & 0.21 & 0.2 & 0.23 & 0.2 & 0.19 & 0.32 \\
microphone & 0.23 & 0.22 & 0.19 & 0.18 & 0.21 & 0.21 \\
microwave & 0.23 & 0.36 & 0.35 & 0.22 & 0.25 & 0.48 \\
motorcycle & 0.27 & 0.12 & 0.22 & 0.24 & 0.28 & 0.34 \\
mug & 0.13 & 0.11 & 0.15 & 0.11 & 0.17 & 0.15 \\
piano & 0.17 & 0.11 & 0.16 & 0.14 & 0.17 & 0.21 \\
pillow & 0.19 & 0.14 & 0.17 & 0.18 & 0.3 & 0.39 \\
pistol & 0.29 & 0.22 & 0.23 & 0.25 & 0.3 & 0.35 \\
pot & 0.19 & 0.15 & 0.19 & 0.14 & 0.16 & 0.25 \\
printer & 0.13 & 0.11 & 0.13 & 0.11 & 0.14 & 0.19 \\
remote & 0.3 & 0.33 & 0.31 & 0.31 & 0.37 & 0.44 \\
rifle & 0.43 & 0.28 & 0.3 & 0.36 & 0.48 & 0.55 \\
rocket & 0.34 & 0.2 & 0.23 & 0.26 & 0.32 & 0.2 \\
skateboard & 0.39 & 0.11 & 0.39 & 0.35 & 0.47 & 0.58 \\
sofa & 0.24 & 0.23 & 0.27 & 0.21 & 0.27 & 0.34 \\
stove & 0.2 & 0.19 & 0.24 & 0.18 & 0.19 & 0.33 \\
table & 0.31 & 0.24 & 0.34 & 0.26 & 0.34 & 0.42 \\
telephone & 0.33 & 0.42 & 0.45 & 0.4 & 0.5 & 0.5 \\
tower & 0.24 & 0.2 & 0.25 & 0.25 & 0.25 & 0.33\\
train & 0.34 & 0.29 & 0.3 & 0.32 & 0.38 & 0.51\\
vessel & 0.28 & 0.19 & 0.22 & 0.23 & 0.29 & 0.35 \\
washer & 0.2 & 0.31 & 0.31 & 0.21 & 0.25 & 0.32 \\
\hline
\end{tabular}
\caption{F-score evaluation (@1\%) in the viewer-centered mode, per-class results.}
\vspace{2mm}
\label{tab:im2p_perclass_results}
\end{center}
\end{table*}

\clearpage

{\small
\bibliographystyle{ieee_fullname}
\clearpage
\bibliography{refs}
}

\end{document}